\documentclass[final,3p,times,twocolumn,authoryear]{elsarticle}

\usepackage{amssymb}
\usepackage{amsmath,soul,color,caption,booktabs}
\usepackage{hyperref}

\definecolor{myred}{rgb}{1.0, 0.0, 0.0}
\definecolor{mygreen}{rgb}{0,0.6,0}
\definecolor{myblue}{rgb}{0.0, 0.0, 1.0}
\definecolor{mylightblue}{rgb}{0.4,0.6,1.0}

\makeatletter
\def\ps@pprintTitle{%
\let\@oddhead\@empty
\let\@evenhead\@empty
\def\@oddfoot{}%
\let\@evenfoot\@oddfoot}

\def\hlinewd#1{%
\noalign{\ifnum0=`}\fi\hrule \@height #1 \futurelet
\reserved@a\@xhline}
\makeatother

\usepackage{lineno}

\journal{arXiv}

\begin{document}

\begin{frontmatter}



\title{First-of-its-kind AI model for bioacoustic detection using a lightweight associative memory Hopfield neural network}



\author[wolves]{Andrew Gascoyne} 
\author[wolves]{Wendy Lomas} 

\affiliation[wolves]{organization={Department of Computing and Mathematical Sciences, University of Wolverhampton},
            addressline={Wulfruna Street}, 
            city={Wolverhampton},
            postcode={WV1 1LY},
            country={UK}}

\begin{abstract}

A growing issue within conservation bioacoustics is the task of analysing the vast amount of data generated from the use of passive acoustic monitoring devices.  In this paper, we present an alternative AI model which has the potential to help alleviate this problem.  Our model formulation addresses the key issues encountered when using current AI models for bioacoustic analysis, namely the: limited training data available; environmental impact, particularly in energy consumption and carbon footprint of training and implementing these models; and associated hardware requirements. The model developed in this work uses associative memory via a transparent, explainable Hopfield neural network to store signals and detect similar signals which can then be used to classify species. Training is rapid ($3$\,ms), as only one representative signal is required for each target sound within a dataset. The model is fast, taking only $5.4$\,s to pre-process and classify all $10384$ publicly available bat recordings, on a standard Apple MacBook Air.  The model is also lightweight with a small memory footprint of $144.09$\,MB of RAM usage.  Hence, the low computational demands make the model ideal for use on a variety of standard personal devices with potential for deployment in the field via edge-processing devices.  It is also competitively accurate, with up to $86\%$ precision on the dataset used to evaluate the model.  In fact, we could not find a single case of disagreement between model and manual identification via expert field guides. Although a dataset of bat echolocation calls was chosen to demo this first-of-its-kind AI model, trained on only two representative calls, the model is not species specific.  In conclusion, we propose an equitable AI model that has the potential to be a game changer for fast, lightweight, sustainable, transparent, explainable and accurate bioacoustic analysis.
\end{abstract}





\end{frontmatter}



\section{Introduction}\label{sec:intro}

The combination of passive acoustic monitoring (PAM), via the use of autonomous recording units (ARUs), and bioacoustic analysis is a cost-effective, non-invasive and sustainable method increasingly used for ecological discovery, monitoring and conservation in most known ecosystems around the globe \citep{2020AbrahamsG,2022Bakker,2023BradferDEJM,2019TeixeiraMR}.  Even when PAM is implemented with the use of careful and consistent guidelines and methods \citep{2023MetcalfAAB,2021PerezT} it still has obvious limitations and does not replace experts working in the field.  With the increase in lower cost ARUs, such as the Apodemus Pippyg, Audiomoth, Song Meter and even custom made devices   \citep{2015Apodemus,2019HillPSD,2017MydlarzSB,2024WA} and the potential of biodegradable sensors \citep{2022SethiKWM}, ecologists and conservationists are now able to leverage new digital and computational technologies for PAM. Given the consequences of rampant biodiversity loss on a warming planet, the vast amount of data collected daily and the limitations of commercial classifiers currently available, there exists an urgent need for new efficient automated analysis tools and software to alleviate the processing bottleneck \citep{2021GoodwinG, 2025KershenbaumABB,2018MacGBBFF,2024McEwanKGB,2022Stowell} without compromising ecological and conservation goals and values \citep{2025Sandbrook}.

Automated techniques to analyse PAM datasets initially involved machine learning via statistical models and ensemble learning and are still used successfully. For example, researchers have used random forest models \citep{2022YohKMA} and many researchers and fieldworkers use Kaleidoscope Pro where, in auto-ID mode, a species-level Hidden Markov Model is created and the results are obtained via clustering analysis \citep{2017WA,2022ManzanoBBSP}. However, deep learning and specifically convolutional neural networks (CNNs) currently dominate academic research on automated bioacoustic classification \citep{2025KershenbaumABB,2024RasmussenSB,2022Stowell}.  CNNs are commonly used for image classification and hence require bioacoustic data to be converted to spectrograms with set dimensions dependent upon the particular architecture of the model, a process which is costly in terms of time, memory and energy. CNNs are considered to be `black boxes' that lack transparency, are computationally expensive and challenging to explain. In ecology and conservation, pretrained models, such as BirdNet \citep{2021KahlWEK}, are popular with architecture most frequently based on the ResNet family of CNN models \citep{2021KahlWEK,2024MacIsaacNAP,2024SalemSSO,2022DufourqBFD}. Although such models show potential and consistent improvement they have yet to be fully proven \citep{2024FunosasBSE,2023Perez}. CNNs are rapidly being superseded by transformer architecture \cite{2020DosovitskiyBKW}, the architecture driving large language models such as ChatGPT, where GPT is the acronym for generative pre-trained transformer. These vision transformer (ViT) models are now slowly being adopted by the bioacoustic research community \citep{2024McEwanKGB,2025FerreiraFMR}. Additionally, and inspired by ViT architecture, a new generation of CNNs, ConvNeXts, has emerged and is producing promising results \citep{2022LiuMWF}. \cite{2025HeinrichRSS} combine a ConvNeXt with a ProtoPNet, a prototypical part network \citep{2019ChenLTB} to produce a more interpretable model for bird sound classification.


All deep learning methods, with their multilayer architecture, are computationally expensive \citep{2021ThompsonGLM}. They require extensive preprocessing of data, up-to-date hardware (CPUs, GPUs and large amounts of RAM) and are usually performed on high performance computing clusters often at Global North institutes \citep{2021KahlWEK,2024MacIsaacNAP}. Not all researchers and few fieldworkers have access to these high performance facilities and expensive hardware. For instance, the single GPU used in \cite{2025HeinrichRSS}, an NVIDIA A100-SXM4-80GB GPU, currently costs £13,800 for the graphics card alone. Training times tend not to be documented in the bioacoustics research literature, although they can be anything from 16 minutes to 2.5 hours per epoch; in \cite{2025HeinrichRSS} each model is trained for 10 epochs.  Furthermore, the environmental impact of the pre-trained models themselves should not be dismissed \citep{2021ThompsonGLM,2020Toews}, as with all industries, use justifies retraining, which in turn uses massive amounts of energy.  For example, training the ResNet-50 on a NVIDIA M40 GPU takes 14 days \citep{2018YouZHDK}. For those researchers looking to estimate their training carbon footprint when using deep learning models, \cite{2023BouzaBL} provide an introduction to energy consumption tracking tools.

Deep learning models require extremely large datasets for pretraining such as ImageNet; however, we cannot assume that the labelling of such datasets is accurate nor that higher capacity CNN models, such as the later ResNet models, will demonstrate better real-world performance than low capacity models as detailed in \cite{2021NorthcuttAM}.  Therefore, layers which have been pretrained will still maintain any legacy/bias learning from the pretrained model.  Additionally, the bioacoustic WAV datasets used to train the final layers are mainly examples of weakly labelled data \citep{2024Planque}, where an entire recording has a single label which is segmented for training. Each segment will then have the same label despite other species or absences being present within the same recording \citep{2024Planque}. Activity detectors are used to partly alleviate this problem \citep{2023GhaniDKK}. A third data issue with all these deep learning methods is that they require balanced datasets which are often unavailable, especially for rare species or specific calls. Data augmentation is having some success \citep{2024MacIsaacNAP} and pretrained transformer models perform well even on smaller training sets \citep{2024McEwanKGB,2025FerreiraFMR}, but all still require significant amounts of labelled data and the use of high performance computing centres for pretraining and transfer learning.  These intensive demands mean that some researchers resort to, and in some cases prefer, manual analysis of sound files and spectrograms, to ensure rare events are discovered and any classification is accurate \citep{2023SzesciorkaMO}, while \cite{2025Sandbrook} warns practitioners of the unintended consequences of the use of AI for conservation.

For researchers looking to make event detections amongst the large numbers of PAM generated files, automated approaches are absolutely invaluable if they are efficient, transparent and sustainable enough to justify their use; ideally, incorporated into edge-processing devices or in active learning approaches of human-computer interaction \citep{2022Stowell}.  The digital age has enabled us to enhance many of our own capacities. In this work, we enhance and augment the listening and recognition capabilities of ecologists and conservationists and the recorders they use in the field today with a new approach of using a neural assosiative memory network that can be trained and deployed on a standard working laptop and used to detect specific bioacoustic events.  The human ear and brain constitute a listening, sound storage and recall device which has definite limitations, namely, processing capacity and speed, and the frequency range within which detections can be made, $20$\,Hz to $20$\,kHz for the human auditory range \citep{1997Brownell, 2004DobieH}.  Digital listening devices and AI can listen to and process at speed a broader range of sounds with varying power thresholds and filters, and as such can significantly augment our own abilities.  Biologically inspired neural associative memory networks, such as those first conceptualised in \citet{1982Hopfield} and used here, simulate the associative memory process by storing patterns which may then be recalled/retrieved from noisy or partial patterns, and are thus examples of content-addressable memory systems. John Hopfield's contribution to AI was acknowledged when he jointly won the Nobel Prize in Physics in October 2024 \citep{2024Nobel}.  Hopfield neural networks (HNNs) have been developed and implemented on a number of tasks such as image recognition and optimisation \citep{1998DaiN,2023LiuGCZ} with further improvements made by incorporating characteristics of chaotic dynamics to overcome the tendency of HNNs to converge to non-optimal solutions \citep{1995ChenA,2024RoddenGNBP}.  In this work, we couple an HNN with augmented `hearing' and demonstrate how this type of neural network can address some of the challenges facing the field of bioacoustics.  The model in this paper can be developed without the need for large training datasets, 'black box' multilayer networks and expensive resource-intensive hardware and training times.  We use an inherently interpretable associative memory neural network model which does not use any image processing techniques.  As this model has not been used for bioacoustic event detection before, we use a public bioacoustic dataset developed to facilitate the research of automated classification techniques \citep{2019Bertran,2019BertranAT}.  As befits the introduction and explanation of a new model to the field, the classification task chosen is relatively simple: to identify the echolocation pulses of two cryptic bat species.  Our model is fast, taking only $5.4$\,seconds to train, pre-process and classify all $10384$ publicly available bat recordings \citep{2019Bertran,2019BertranAT}, on a standard Apple MacBook Air. The model is also lightweight, i.e., it has a small memory footprint of $144.09$\,MB of RAM usage.  These low computational demands make the model ideal for use on a variety of standard personal devices with potential for deployment in the field via edge-processing devices.

This paper is organised as follows.  Firstly, in section \ref{sec:material} we outline the dataset used to evaluate our model including its source, structure and relevance for AI model development.  In section \ref{sec:theory} we outline how the model is developed from the underlying theory of Hopfield networks and how patterns are stored in its memory, to the augmentation of hearing onto the model via the fast Fourier transform.  In section \ref{sec:results} we present the results and performance metrics of the model and discuss these results in section \ref{sec:disc} highlighting the key issues and consequences for this model application.

\section{Material and methods}\label{sec:material}

Here we principally explain the dataset used for model development and testing. Bats were chosen as the sound source for this study. Bats are protected in Europe and the UK, and considered to be indicators of biodiversity \citep{2003CattoCAL}. All UK species are nocturnal and therefore bat surveyors are partially dependent upon acoustic data to survey and monitor their populations. Furthermore, there is a long tradition of studying bat echolocation pulses since the pioneering work of Griffin and Pierce in the 1930s \citep{1938PierceG}. Additionally, because the majority of bat vocalisations and echolocation calls are in the ultrasonic frequency range - that is above the frequency at which humans can hear - fieldworkers use a combination of heterodyne (where signals are shifted into the audible frequency range), time-expansion techniques (where the sound is slowed down to the audible range), and spectrograms (visual representations of the signals) to manually identify species. Unsurprisingly, these manual techniques run into problems in large surveys with many thousands of hours of data to slow down and analyse.

The split dataset used in this work was created by \cite{2019BertranAT}, the authors of this paper had two problems in mind:
\begin{itemize}
    \item Identifying two cryptic, or morphologically similar, bat species - Pipistrellus pipistrellus, or the Common pipstrelle (PIPI), and Pipstrellus pygmaeus or Soprano pipistrelle (PIPY) - species which were not considered to be distinct until the late 1990s \citep{1999BarlowJ}.
    \item Creating a dataset suited to training AI models in order to identify these two species in urban environments where man-made sources of ultrasonic sounds are also present.
\end{itemize} 

The signals were recorded by Elena Tena, a co-author of \cite{2019BertranAT}, in an Iberian forest in the Guadarrama Mountains between 2016 and 2018; an area with very little man-made sonic pollution. Echo Meter Touch Pro 1 bat detectors were used to capture the echolocation sequences and filtered using Kaleidoscope (Wildlife Acoustics, Inc., USA). Analysis and labelling were completed via commercial software, BatSound 4 (Pettersson Elektronik AB, Upsala Sweden), and expert manual confirmation \citep{2019BertranAT}. The creators then split the sound files into single labelled echolocation pulses (milliseconds in length, varying from less than $0.5$\,ms to nearly $30$\,ms) and silences. The downloadable dataset contains $4916$ PIPI fragments, or echolocation pulses, $5064$ PIPY fragments and $12187$ fragments of silence.  These silences include some longer files (up to $400$\,ms) containing no echolocation pulses, and also some much shorter files containing inter-echolocation pulse intervals.  We decided to use a subset of the dataset with all PIPI and PIPY fragments but with a reduced number of silences ($404$).  Filtering out silences is a fairly simple task of pre-processing the data before passing to the model.  We did not want this to skew the model metrics in our favour or to distract from the auto-detection AI task so we only used the longer silence files when developing our prototype. Therefore, our initial testing dataset contained the remaining $10384$ sound files.

Much of the research into AI models for bioacoustics today involves the conversion of a sound file to an image before passing to a CNN for classification \citep{2025KershenbaumABB,2024RasmussenSB,2022Stowell}. However, computationally this is a costly process. While a fast fourier transform (FFT) efficiently discretises the signal from the time to the frequency domain, reassembling this information into the time domain to create a spectrogram (a heatmap showing frequency, power and time information) is costly which we will discuss in detail in section \ref{sec:disc}.  The model developed here will only use the FFT of the signal and not the spectrogram thus reducing the computational time of the algorithm to process the signals.  We should also note here that the short split signals that make up this dataset are not a requirement for the model and are merely a feature of the dataset.  Longer signals can be passed to the model for bioacoustic event detection if required.

\section{Theory and calculation}\label{sec:theory}
In this section we describe the model formulation.  We suggest a Hopfield network model and combine this with signal processing techniques in order to train the model and hence identify bioacoustic signals.

\begin{figure}[ht]
\centering
\includegraphics[width=0.4\textwidth]{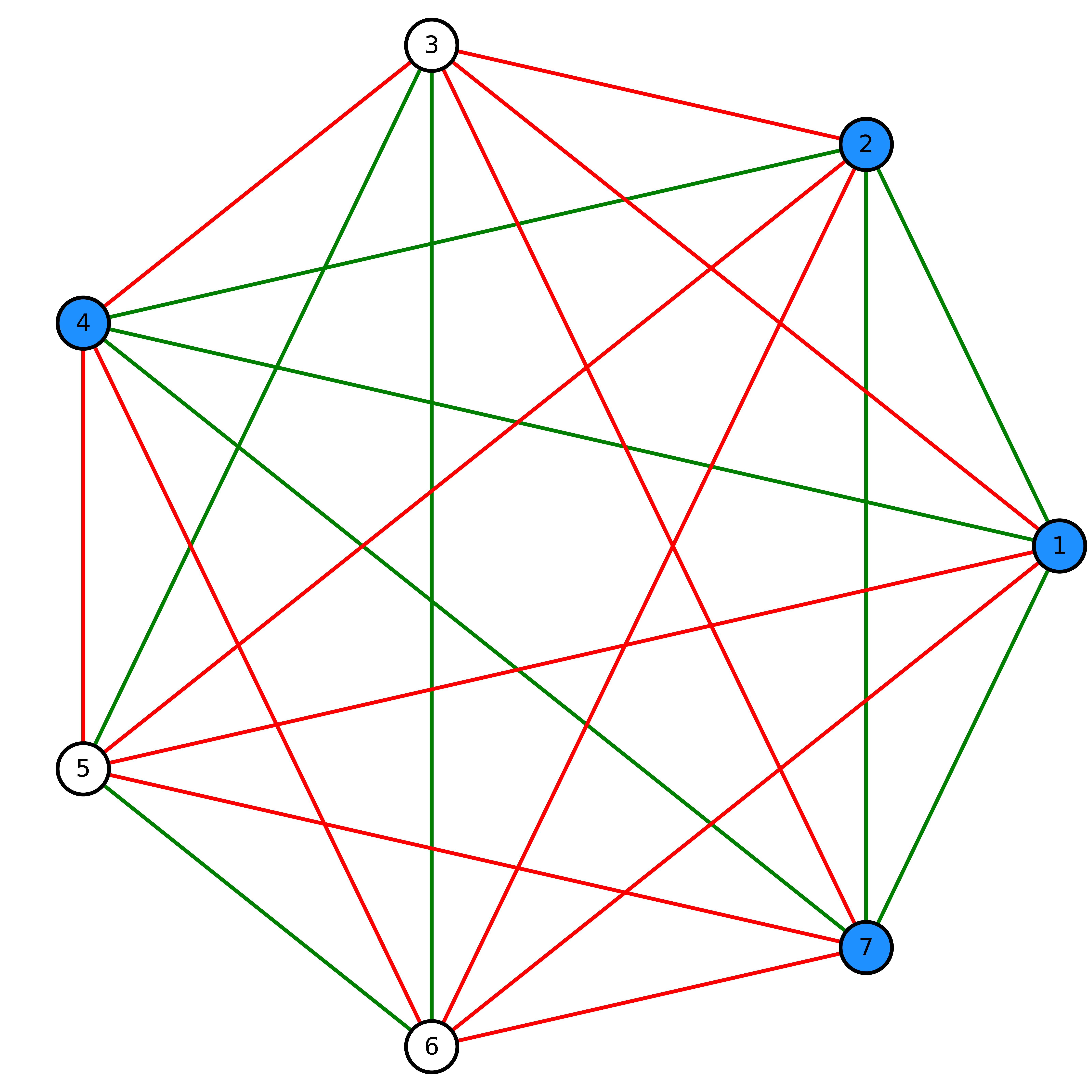}
\caption{Network diagram for the example discussed. 
Blue vertices represent activated/firing neurons, whereas white vertices represent dormant/non-firing neurons.  Green edges represent connections between neurons with the same state whereas red edges represent neuronal connections with opposite states.}\label{fig:netex}
\end{figure}

\begin{figure*}[ht]
\centering
\includegraphics[width=0.95\textwidth]{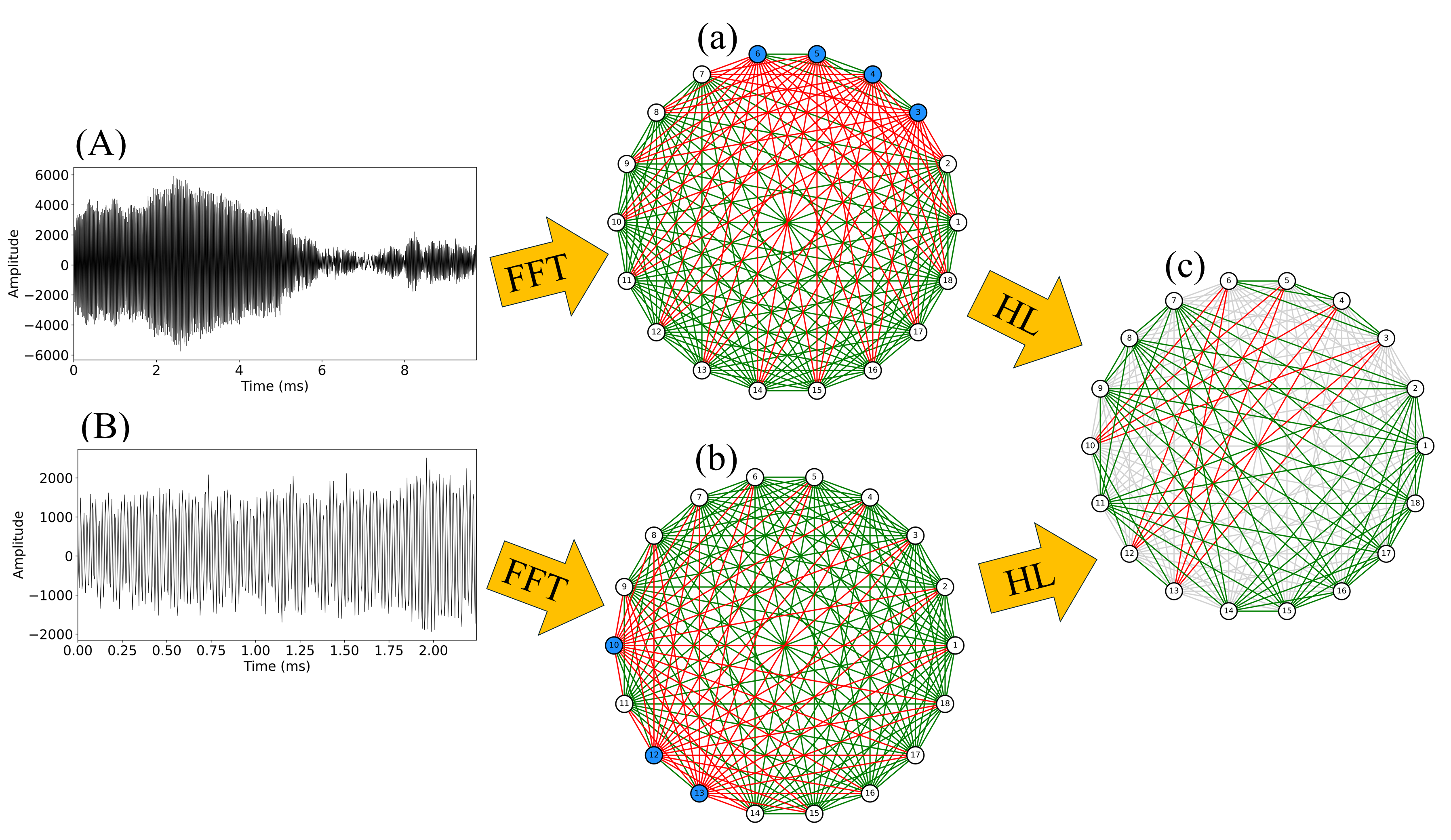}
\caption{Fast Fourier Transform (FFT) is applied to the signals (A) and (B) which are the echolocation calls for each species PIPI and PIPY respectively.  These are then used to activate the network, (a) and (b) respectively, where neurons which are fired are indicated in blue.  In order for both of the echolocation calls to be stored in the network memory we combine the two network activations (a) and (b) using Hebbian learning \eqref{eqn:weights} to form the trained network model (c), which is now ready for activation.
}\label{fig:mflow}
\end{figure*}

\subsection{Hopfield Network Model}\label{sec:Hop}
Hopfield networks are recurrent neural networks with associative memory patterns, and unlike the majority of neural network architectures used, such as feed-forward, Hopfield networks are fully connected \citep{1984Hopfield}.  Hence, there are no separate input or output neurons.  Instead the mean internal potential of the neuron is converted into a firing rate output of the neuron causing the network state to evolve with time.  The classical Hopfield network for continuous variables can be described by the dynamical equations \citep{1985HopfieldT}:
\begin{equation}\label{eqn:update}
\frac{d y_i}{dt}=-\frac{y_i}{\tau}+\sum_i\sum_j w_{ij}x_j+I_i
\end{equation}
where $\tau$ is a positive constant (resistance-capacitance); $y_i(t)$ is the internal state of the $i$-th neuron at time $t$; $x_i(t)$ is the activity of the $i$-th neuron at time $t$, $w_{ij}$ is the connection weight from neuron $j$ to neuron $i$; $I_i$ is the input bias of neuron $i$.  In order to ensure convergence of the system to stable states where the output neurons, $x_j$, all remain constant we ensure the network connection weights, $w_{ij}$ are symmetric, i.e., $w_{ij}=w_{ji}$.  If we also constrain the neurons to take on binary outputs, known as the high-gain limit, and there are no self-connections, $w_{ii}=0$, the system Lyapunov function, also known as ``energy'', is given by:
\begin{equation}
E=-\frac{1}{2}\sum_i\sum_j w_{ij}x_ix_j-\sum_i I_ix_i
\end{equation}
as described in \citet{1985HopfieldT}.  The stable states which the network converges to are the minima of this energy function.  The state space of the network is the interior of the $N$-dimensional hypercube defined by $x_i=-1$ or $1$.  In our formulation the minima occur at the corners of this hypercube.

\begin{figure*}[ht]
\centering
\includegraphics[width=0.99\textwidth]{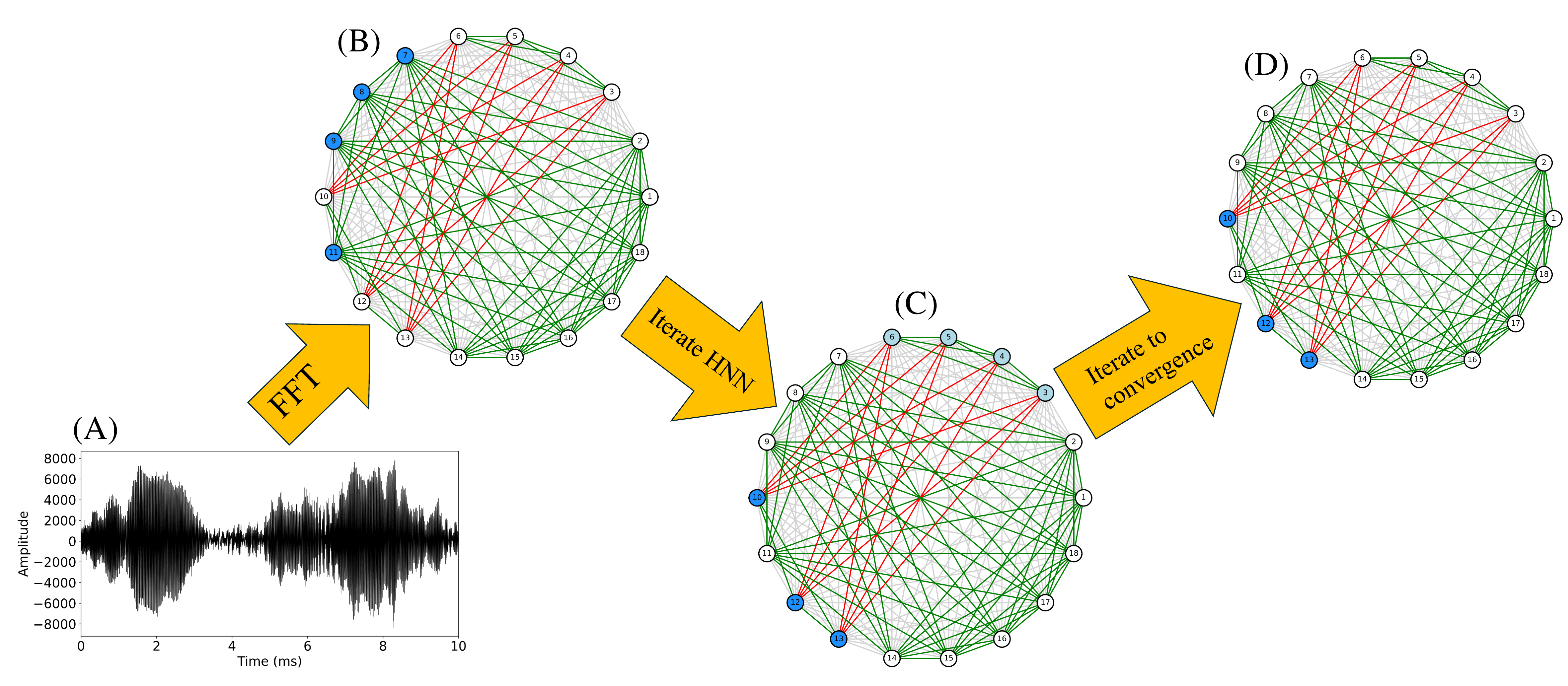}
\caption{Fast Fourier Transform (FFT) is applied to the input signal (A) which may or may not contain echolocation calls for either species PIPI or PIPY.  The FFT of the signal is then used to activate the trained network (B), where fired ($x_i=+1$) and non-fired ($x_i=-1$) neurons are blue and white respectively.  The network then updates using equation \ref{eqn:update} to form the next iteration of the HNN (C).  Observe that in the second iteration (C), neutral neurons ($x_i=0$) are produced in light blue which are neither fired nor dormant.  We continue to iterate the network until it converges which results in the final state of the network (D).  We then compare the final state to the stored signals in the network memory, retrieval states (a) and (b) in figure \ref{fig:mflow}.  Hence, label this signal as either PIPI or PIPY if the final state is a retrieval state or label as UnID if the network converges to a spurious state.
}\label{fig:mflowP}
\end{figure*}

The Hopfield network is of particular use to optimisation problems due to the guaranteed convergence of the network to local minima.  In order to construct and apply a Hopfield network to a particular optimisation problem we must define the network weights $w_{ij}$.  One method for assigning connection weights is to invoke Hebb's rule or Hebbian learning \citep{1949Hebb}, the principle being ``neurons that fire together, wire together''.  Therefore, neurons that are active simultaneously become associated such that future activity in one will affect the activity of its associate neurons.  In contrast, neurons that are not linked in this way, become less connected or associated.  As an example for a network with $N=7$ neurons we will store a single network configuration given by $\mathbf{X}^T=[1,1,-1,1,-1,-1,1]$.  The weights, $\mathbf{W}$, are computed via Hebb's rule as follows:
\begin{equation}
\mathbf{W}=\mathbf{X}\cdot\mathbf{X}^T=\begin{bmatrix}
  1 & 1 & -1 & 1 & -1 & -1 & 1\\
  1 & 1 & -1 & 1 & -1 & -1 & 1\\
  -1 & -1 & 1 & -1 & 1 & 1 & -1\\
  1 & 1 & -1 & 1 & -1 & -1 & 1\\
  -1 & -1 & 1 & -1 & 1 & 1 & -1\\
  -1 & -1 & 1 & -1 & 1 & 1 & -1\\
  1 & 1 & -1 & 1 & -1 & -1 & 1\\
\end{bmatrix}
\end{equation}
$\mathbf{X}$ in this example is known as a retrieval state \citep{1991Hertz} and we represent this states network configuration in figure \ref{fig:netex}.  In the case where multiple configurations are stored into the network we use the following formula to calculate the network weights:
\begin{equation}\label{eqn:weights}
\mathbf{W}=\frac{1}{N}\sum_{k=1}^p\mathbf{X}^k\cdot\mathbf{X}^k
\end{equation}
where $p$ is the number of configurations stored in the network.  \citet{1982Hopfield} showed that ``about $0.15N$ states can be simultaneously remembered before error in recall is severe'', hence we must be mindful of this constraint when constructing network architectures for successful memory recall.  Also, it is possible for the Hopfield network to converge to states other than the retrieval states such as reversed states, mixture states and spin glass states \citep{1991Hertz}, these are known as spurious states.

\subsection{Hopfield Network as a Bioacoustic Indentifier}
We want our model network to be able to identify bat species based on their echolocation pulses, therefore we must represent these sound signals in the network.  The issue we encounter is that, at their essence, signals are continuous and yet we wish to represent the signal within the network which is discrete.  Therefore we take inspiration from biology, specifically the ear and brain.  The ear and brain constitute a discretising listening, storage and recall device whereby continuous sound waves enter the ear as pressure perturbations which transduce in the middle ear into mechanical energy.  These waves cause the tympanic membrane to vibrate, setting the ossicles (malleus, incus and stapes) in motion and in turn causing the oval window at the cochlea entrance to vibrate.  This vibration causes the travelling wave to pass through the inner ear. The fluid filled cochlea is set in motion and a selection of the thousands of tiny auditory hair cells bend as the wave passes through the spiral, ultimately causing another transduction to electro-chemical neural impulses which transport information about the signal to the brain stem.  The auditory hair cells act as biotransducers, discretising a continuous travelling wave into narrow bands of frequencies, characteristic of the location of the hair cell, high frequencies being detected first as the transformed waves enter the cochlea \citep{1997Brownell,2004DobieH}.  The coded discretised information about the sound waves is transmitted to the brain, relaying frequency, power, and temporal characteristics, and enabling a 3-dimensional acoustic representation of the world to be recreated.

The brain part of our model is represented by our Hopfield network (described in section \ref{sec:Hop}), we just to need represent the ear aspect.  For this we will use signal processing in order to digitise the signal ready for the network.  The dataset which will be used to test the model is composed of 10384 1-dimensional wave files, hence the Fast Fourier transform (FFT) is a suitable algorithm to digitise such signals.  Firstly we must train the model by selecting echolocation calls for each species from the dataset which are typical of those made by the respective species PIPI and PIPY.  The FFT will convert the signal from the time domain into the frequency domain and peak frequencies are selected from each signal in order to activate the network.  Using Hebbian learning these network configurations are stored in the network as retrieval states as described in figure \ref{fig:mflow}.  With our Hopfield network model constructed we use the remaining wave files in the dataset to test our model, which will attempt to identify them.  In figure \ref{fig:mflowP} we describe how signals are sent to the trained model (see figure \ref{fig:mflow}) for prediction.  An advantage of this algorithm is that we can easily observe how the network configuration evolves from one iteration to the next and therefore it is clear to understand how the model makes its predictions.  Hence, the algorithm described here produces explainable AI models and therefore users can trust the results and outputs produced.

As outlined in section \ref{sec:Hop} it is possible the network does not converge to the retrieval states which correspond to the PIPI and PIPY echolocation calls the network was trained on, i.e., converge to spurious states of the network.  This is in fact an advantage of this model since calls which are not identified as either PIPI or PIPY are labelled unidentified (UnID) by the model and can therefore be investigated separately.

\section{Results}\label{sec:results}%
In this section we will present the results of the model (see section \ref{sec:theory} on model construction) on the \citet{2019BertranAT} dataset (see section \ref{sec:material} for information on the dataset).  We try two versions of the model based on our insights about the dataset:  Model $1$ is tested on the dataset after silences have been filtered out.  Model $2$ is tested on the dataset after silences have been filtered out and also files of echolocation pulses with $\text{F}_{\text{maxE}}$ between $49$ and $51$\,kHz removed \citep{2021Russ,2018AughneyRL,2003CattoCAL}.  Silences were filtered out automatically by determining whether the signal had any peak frequencies above a tunable tolerance before passing the remaining signals to the model.  The algorithm was successful in identifying all silences before passing the remaining $8476$ files to the trained Hopfield network model for identification.  Hence, the performance metrics only relate to the identification of PIPI and PIPY bat species and not on identifying the silences within the dataset, since this would over-inflate the metrics and give us an incorrect impression of performance with regard to identifying the two bat species based on echolocation calls.

\begin{table*}[ht]
\begin{center}\normalsize
 \begin{tabular}{l|cccc|cccc}
   \toprule
   Class & \multicolumn{4}{c|}{Model 1} & \multicolumn{4}{c}{Model 2} \\
   \cmidrule(lr){2-5}\cmidrule(lr){6-9}
    & Precision & Recall & F1 & Support & Precision & Recall & F1 & Support \\
   \midrule
   PIPI & 0.75 & 0.68 & 0.72 & 4193 & 0.79& 0.84& 0.81& 2249 \\
   PIPY & 0.72 & 0.76 & 0.74 & 4283 & 0.86& 0.77& 0.81& 2550 \\
   \midrule\midrule
   & \multicolumn{3}{c|}{Overall Accuracy:  $0.72$} & $8476$ & \multicolumn{3}{c|}{Overall Accuracy:  $0.80$} & 4799 \\
   \bottomrule
 \end{tabular}
\caption{Classification reports for model $1$ and $2$.}
\label{tab:CR}
\end{center}
\end{table*}

\begin{figure}[ht]
 \centering
 \includegraphics[width=0.49\textwidth]{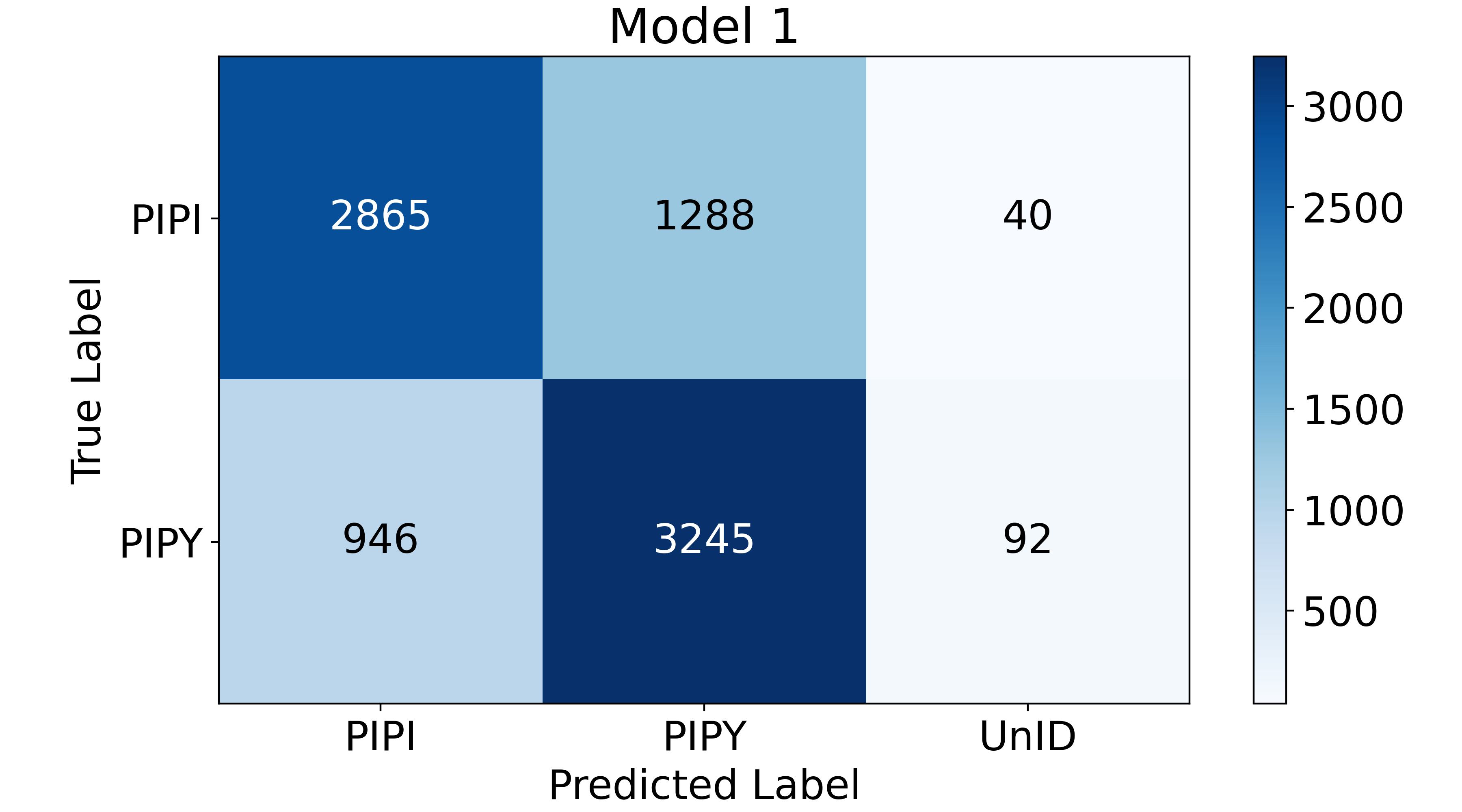}
 \includegraphics[width=0.49\textwidth]{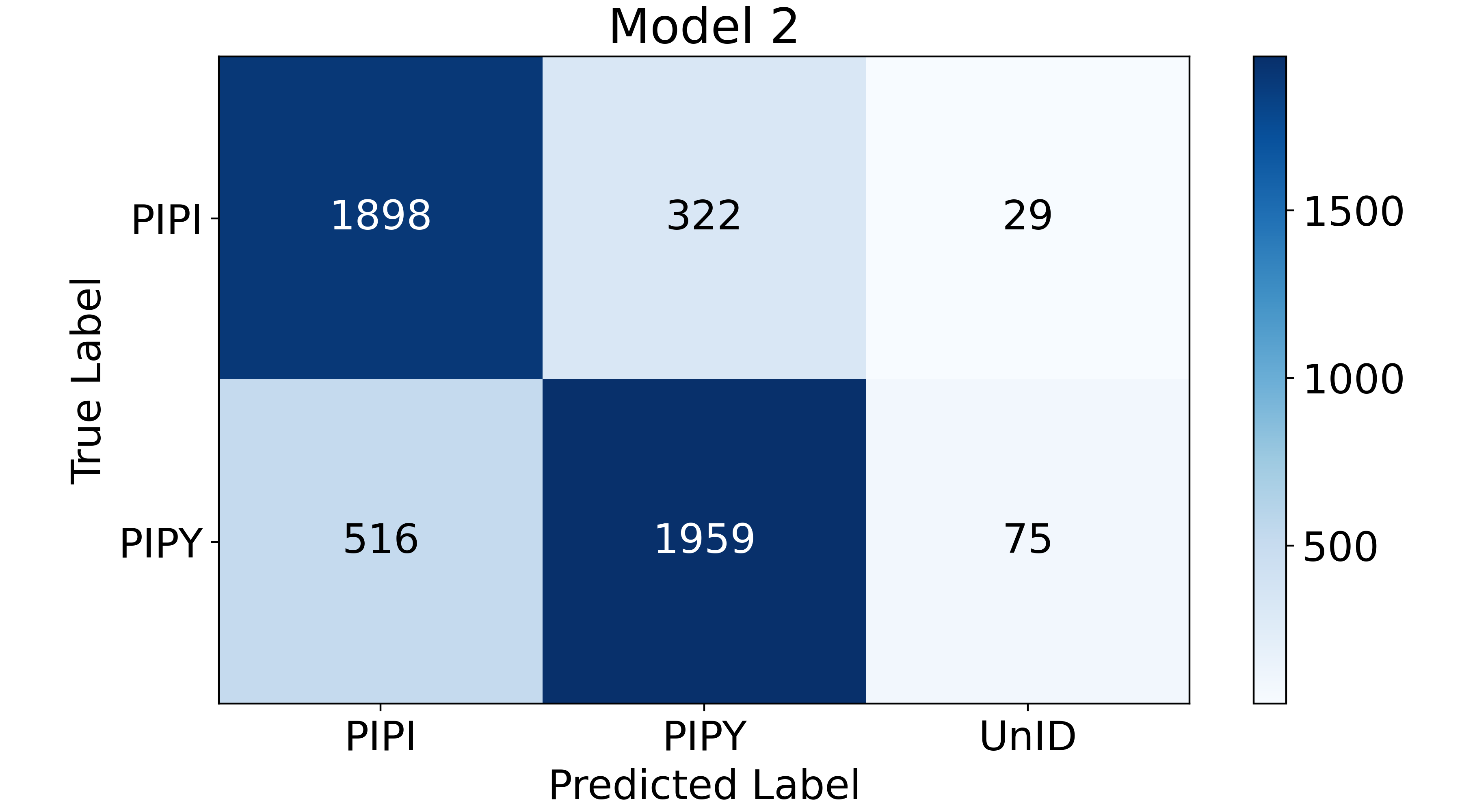}
 \caption{Confusion matrices for model $1$ and $2$.}
 \label{fig:CM}
 \end{figure}

In table \ref{tab:CR} and figure \ref{fig:CM} we present the performance metrics and confusion matrix for both model $1$ and $2$.   For model $1$ we observe an overall accuracy of $0.72$ with the highest metric being the $0.76$ recall score for the PIPY species which is very promising at this stage of model development.  For model $2$ there is an increase in all performance metrics with an overall accuracy of $0.80$ with the highest performance metric of $0.86$ precision score for the PIPY species.  For species surveying purposes, recall is of relevance to give accurate estimates of population size with model $2$ showing better performance for both PIPI and PIPY at $0.84$ and $0.77$ compared to model $1$.  For other bioacoustic monitoring purposes, such as species specific interventions, precision would be the relevant metric with model $2$ showing better performance for both PIPI and PIPY at $0.79$ and $0.86$ compared to model $1$.  The F1 score, the harmonic mean of precision and recall, shows better performance for Model $2$ with consistent scores for both PIPI and PIPY detection.  The model performance is comparable to both commercial software \citep{2022MarchalFA, 2022TabakMRLB} and the models developed by the dataset creators \citep{2019BertranAT}.  We will discuss the aspects of the data the model does not predict so well in section \ref{sec:disc}.

The confusion matrix, in figure \ref{fig:CM}, gives us a breakdown of the model prediction compared to the true label given in the dataset.  Note here we have an extra prediction label UnID, which the model returns when the neural network converges to a spurious state not associated with the training signals, as discussed in section \ref{sec:theory}.  Essentially, for the signals labelled UnID, the model does not associate the signal with either of the PIPI or PIPY signals it was trained on.  There are a total of $132$ signals in model $1$ and $104$ signals in model $2$ which are not identified as either PIPI or PIPY and hence need further investigation which we will discuss in section \ref{sec:disc}.  We observe a marked decrease in the number of mislabelled PIPI signals predicted as PIPY by the model, from $1288$ in model $1$ to $322$ in model $2$.  This infers that there are a large number of PIPI labelled signals in the dataset for which $\text{F}_{\text{maxE}}$ is between $49$ and $51$\,kHz, which were removed from the dataset when testing model $2$ and hence improve the performance of model $2$ compared with model $1$.  This is also true for the PIPY labelled signals in the dataset which where predicted as PIPI by the models.

Not only did model $2$ prove to be competitively accurate it also had low computational demands.  It was developed and tested on a standard working laptop, an Apple MacBook Air. The model is fast, taking on average just $3$\,milliseconds to train and only $5.4$\,seconds to pre-process and classify all $10384$ publicly available bat recordings \citep{2019Bertran,2019BertranAT}. The model is also lightweight, i.e., it has a small average memory footprint of $144.09$\,MB of RAM usage.  To gain these average performance figures model $2$ was run five times in order to calculate the mean time taken and memory footprint. The laptop used was in normal use with other applications and tabs open and active.

\section{Discussion}\label{sec:disc}
In this section we will discuss the results of the models, focusing on where the model did not agree with the dataset labelling of \cite{2019BertranAT}.  These are given by the results in the four misclassified sets: PIPY detected as PIPI, PIPI detected as PIPY, PIPI detected as UnID and PIPY detected as UnID. The fully connected network architecture and iterative convergence process allows for interpretable classifications, as detailed in this section and described in figure \ref{fig:mflowP}, fulfilling the aim of being both transparent and explainable. We then go on to discuss further the computational performance of the model.

\subsection{Dataset Limitations}
Upon analysing the results of the two models in section \ref{sec:results} we notice three key issues which require discussion:
\begin{enumerate}
\item\label{pt:50kHz} A large number of files ($3677$) had been labelled either PIPI or PIPY in the dataset even though the literature suggests that these signals should not be associated with either species \citep{2021Russ,2003CattoCAL}.
\item\label{pt:UnID} Both models $1$ and $2$ returned $132$ and $104$ files respectively as unidentified results. 
\item\label{pt:misclass} Although a small percentage overall, there were a total of $838$ mislabelled predictions made by model $2$; either PIPI identified as PIPY or PIPY identified as PIPI. 
\end{enumerate}

\begin{figure*}[ht]
\centering
\includegraphics[width=0.95\textwidth]{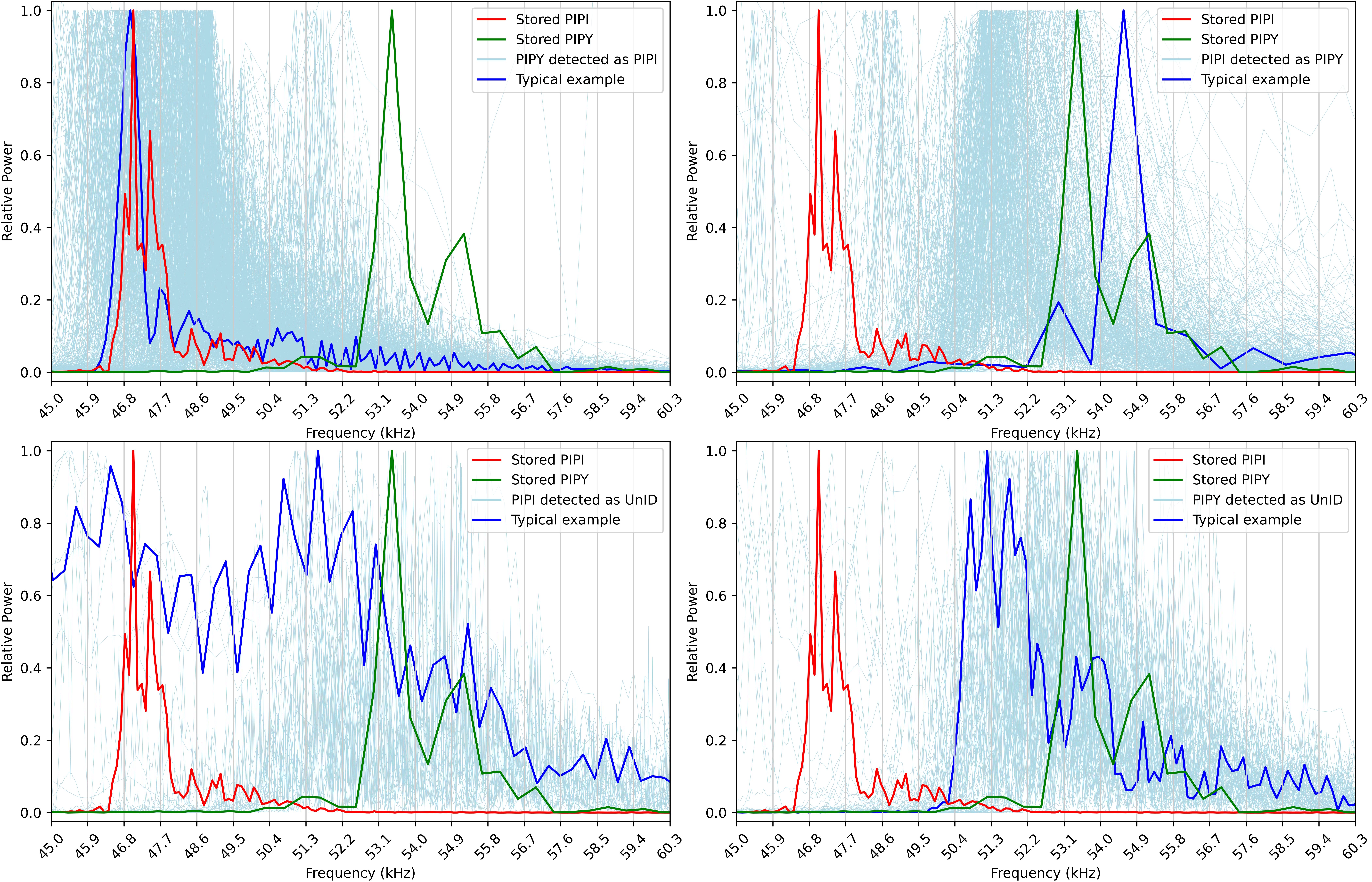}
\caption{Full power spectral density plots comparing the species PIPI (\textcolor{myred}{\bf red}) and PIPY (\textcolor{mygreen}{\bf green}), which were stored in the neural network memory as retrieval states, with all of the cases of misclassified signals plotted in \textcolor{mylightblue}{\bf light blue} for the indicated set.  A typical example of a misclassified signal within the indicated set is shown in \textcolor{myblue}{\bf blue}.}
\label{fig:misclass}
\end{figure*}

We have tried to account for point \ref{pt:50kHz} by using two versions of the model:  model $1$ and $2$ evaluated on the dataset with and without these $3677$ files included (see section \ref{sec:results}) to compare performance.  Not surprisingly we do see an improvement in the performance in model $2$ compared to model $1$.  It is argued that these signals should not be considered echolocation calls from either PIPI or PIPY bat species and therefore we cannot expect the model to identify them.

Referring to point \ref{pt:UnID}, the model developed here has a third class, `UnID' (not present in the original dataset), which the model will return if the network converges to a spurious state not associated with either of the PIPI or PIPY echolocation calls stored in the network memory (retrieval states).  These spurious states returned by the network are all reversed states, therefore not characteristic of either retrieval state.  This is useful since the model will notify us when it does not recognise a signal as any stored within its memory and therefore we can immediately be made aware of unexpected and problematic recordings or potential model problems. This third class proved invaluable for initial result evaluation and understanding dataset limitations, and leveraged the network's spurious states to express uncertainty.

In order to address point \ref{pt:misclass} (and point \ref{pt:UnID}), and better understand how the model misclassified signals we plot the power spectral density of each of the four misclassified sets along with the original PIPI and PIPY species in figure \ref{fig:misclass}.  Using expert field guides and reports \citep{2021Russ,2018AughneyRL,2003CattoCAL} we can manually identify the calls to determine if we would classify these any differently to the model.  Generally PIPI are considered identifiable if echolocating within the range of our model but below $48$\,kHz, and PIPY identifiable if echolocation pulses are above $52$\,kHz.  In the top panels of figure \ref{fig:misclass} we would argue that our manual identification would agree with that of the model for the vast majority of signals and therefore disagree with the dataset labelling of these signals.  In the bottom panels of figure \ref{fig:misclass}, where the model returned `UnID', these signals typically have frequencies either across the entire interval or inside the $49$ to $51$\,kHz region.  Hence, we would argue that the pattern of frequencies present in the power spectra is not distinct enough for ID as either pipistrelle species and these pulses were almost certainly not assigned correctly in the dataset.  

We conclude that the overwhelming majority of the signals not detected correctly were either mislabelled in the original dataset or labelled despite expert consensus \citep{2021Russ,2018AughneyRL,2003CattoCAL}; in isolation as single pulses these signals should not be considered indicative echolocation calls for these species.  The call itself may not be species indicative for a variety of reasons:  pipistrelles modifying their echolocation calls based on interaction with other bats, foraging or environmental factors.  However, examination of the original file before the signals were split into single pulses, confirmed our mislabelling hypothesis.

\subsection{Model Performance and Comparison}
Model $2$ proves to be competitively accurate with an overall accuracy of $80$\% as detailed in section \ref{sec:results} and also more importantly here has low computational demands.  It was developed and tested on a standard working laptop, an Apple MacBook Air. The model proved to be extremely fast, taking on average just $3$\,milliseconds to train and only $5.4$\,seconds to pre-process and classify all $10384$ publicly available bat recordings \citep{2019Bertran,2019BertranAT}. Although the dataset creators \cite{2019BertranAT} did develop a CNN with accuracy of $74$\% for bat detection, computational demands were not discussed in their paper. On the laptop running our models we ran a short experiment to determine how long it would take to convert $10384$ sound files to spectrograms. Initially this was run through a Python IDE which crashed on every attempt. When run through the terminal, with low resolution spectrogram settings, the best we could manage was $6.5$\,minutes to convert the files. In order to make these images suitable for a CNN, further processing would be required due to the irregular lengths of the WAV files within the dataset \citep{2019Bertran,2019BertranAT}. Creating these images took over $72$ times longer than our model takes to train, pre-process and classify all the files; recall our model processes the WAV files directly without the need to create spectrograms. Simply creating the spectrogram images takes $130000$ times longer than training our model, and indicates how costly the conversion to spectrograms is in CNN models. Not to mention the $14$ days to train CNNs from scratch when taking into account pretraining \citep{2018YouZHDK}. Furthermore, local storage capacity is impacted as even these low resolution images result in over twice the amount of memory being used for storage prior to processing and classification. Our model is also lightweight, i.e., it has a small average memory footprint of $144.09$\,MB of RAM usage.  We should emphasise again here that this model can be built and run without requiring any access to high performance computing resources.

\section{Conclusion}

The model developed demonstrates the effectiveness of using associative memory as the training mechanism in order to detect bioacoustic events.  This is advantageous due to the minimal training sets required which results in a fast and lightweight model trained in just $3$\,milliseconds, pre-processing and classifying all $10384$ publicly available bat recordings in $5.4$\,seconds.  The model is lightweight since it is deployed on a standard working laptop (Apple MacBook Air) and has a small average memory footprint ($144.09$\,MB of RAM usage). This is in stark contrast to the CNN approach more commonly used which requires vast amounts of labelled training data and signifcant computational resources (GPUs, high performance computing resources etc.).  These types of models are ideal for bioacoustic monitoring given the vast amounts of unlabelled data collected (limited training data), importance of rare species detection and limited resources in the field.  We have shown that the model is competitively accurate, assisting in essential problem solving within the field of conservation. The model's low energy, time and hardware requirements mean that its use is sustainable and fits well with conservation, ecology and equitable AI long term goals.

A further advantage of using an HNN is that the model is inherently interpretable. We can dig into each classification and discover exactly how it was made, see figure \ref{fig:mflowP}. This transparency and explainability was particularly advantageous, when evaluating the model performance.  At this point we became aware of the limitations of the dataset: principally the inclusion and labelling of calls which are widely agreed to be inappropriate for classification; and the mislabelling of a significant proportion of files.  Hence, although the dataset and the classification task, were chosen in order to best introduce and demonstrate a novel algorithm to building AI models in the field of bioacoustics, the performance metrics do not tell the whole story.  In fact, with further analysis of the results, it was hard to find any examples of disagreement between model and manual identification using expert field guides.  It should be stressed that the model was not designed to classify species but to detect similar signals within its memory.  We train the model on species specific echolocation calls and from this we infer the species of bat. Hence, models can therefore be constructed to detect different species or particular vocalisations.

Biologically inspired mathematical models, such as these associative memory HNNs, have great potential to assist with modern problems, but as we have outlined here the current emphasis amongst Global North researchers is on large training datasets, heavyweight convolutional neural networks and large language models. We suggest that brain-inspired lightweight associative memory models offer a sustainable and equitable way forward as we move into an era where reducing human environmental impact is critical to the survival of all species on a warming planet.



\section{Acknowledgements}
We thank the OpenBright Foundation and trustee Elizabeth Molyneux for their support and funding.  We also thank the University of Wolverhampton for the Invest to Grow PhD Studentship funding.  The authors would also like to thank Catherine Povey (Just Mammals Ltd) for her expertise and useful discussions.





  \bibliographystyle{elsarticle-harv} 
  \bibliography{GascoyneLomas_rev}

\def\cprime{$'$}
\begin{thebibliography}{61}
\expandafter\ifx\csname natexlab\endcsname\relax\def\natexlab#1{#1}\fi
\providecommand{\url}[1]{\texttt{#1}}
\providecommand{\href}[2]{#2}
\providecommand{\path}[1]{#1}
\providecommand{\DOIprefix}{doi:}
\providecommand{\ArXivprefix}{arXiv:}
\providecommand{\URLprefix}{URL: }
\providecommand{\Pubmedprefix}{pmid:}
\providecommand{\doi}[1]{\href{http://dx.doi.org/#1}{\path{#1}}}
\providecommand{\Pubmed}[1]{\href{pmid:#1}{\path{#1}}}
\providecommand{\bibinfo}[2]{#2}
\ifx\xfnm\relax \def\xfnm[#1]{\unskip,\space#1}\fi
\bibitem[{Abrahams and Geary(2020)}]{2020AbrahamsG}
\bibinfo{author}{Abrahams, C.}, \bibinfo{author}{Geary, M.},
  \bibinfo{year}{2020}.
\newblock \bibinfo{title}{Combining bioacoustics and occupancy modelling for
  improved monitoring of rare breeding bird populations}.
\newblock \bibinfo{journal}{Ecological Indicators} \bibinfo{volume}{112}.
\newblock \DOIprefix\doi{10.1016/j.ecolind.2020.106131}.
\bibitem[{{Apodemus Field Equipment}(2025)}]{2015Apodemus}
\bibinfo{author}{{Apodemus Field Equipment}}, \bibinfo{year}{2025}.
\newblock \bibinfo{title}{Apodemus pippyg}.
\newblock
  \bibinfo{howpublished}{https://www.apodemus.eu/en/Apodemus-Pippyg/00127}.
\newblock \bibinfo{note}{Accessed: 2025-04-04}.
\bibitem[{Aughney et~al.(2018)Aughney, Roche and Langton}]{2018AughneyRL}
\bibinfo{author}{Aughney, T.}, \bibinfo{author}{Roche, N.},
  \bibinfo{author}{Langton, S.}, \bibinfo{year}{2018}.
\newblock \bibinfo{title}{The Irish Bat Monitoring Programme 2015-2017}.
\newblock \bibinfo{organization}{Irish Wildlife Manuals, No. 103, National
  Parks and Wildlife Service}. \bibinfo{address}{Department of Culture,
  Heritage and the Gaeltacht, Ireland}.
\bibitem[{Bakker(2022)}]{2022Bakker}
\bibinfo{author}{Bakker, K.}, \bibinfo{year}{2022}.
\newblock \bibinfo{title}{The Sounds of Life: How Digital Technology is
  Bringing Us Closer to the Worlds of Animals and Plants}.
\newblock \bibinfo{publisher}{Princeton, NY: Princeton University Press}.
\bibitem[{Barlow and Jones(1999)}]{1999BarlowJ}
\bibinfo{author}{Barlow, K.}, \bibinfo{author}{Jones, G.},
  \bibinfo{year}{1999}.
\newblock \bibinfo{title}{Roosts, echolocation calls and wing morphology of two
  phonic types of pipistrellus pipistrellus}.
\newblock \bibinfo{journal}{International Journal of Mammalian Biology}
  \bibinfo{volume}{64}, \bibinfo{pages}{257--268}.
\bibitem[{Bertran et~al.(2019)Bertran, Alsina-Pagès and Tena}]{2019BertranAT}
\bibinfo{author}{Bertran, M.}, \bibinfo{author}{Alsina-Pagès, R.},
  \bibinfo{author}{Tena, E.}, \bibinfo{year}{2019}.
\newblock \bibinfo{title}{Pipistrellus pipistrellus and pipistrellus pygmaeus
  in the iberian peninsula: An annotated segmented dataset and a proof of
  concept of a classifier in a real environment}.
\newblock \bibinfo{journal}{Applied Sciences} \bibinfo{volume}{9}.
\newblock \DOIprefix\doi{10.3390/app9173467}.
\bibitem[{Bertran~Ferrer(2019)}]{2019Bertran}
\bibinfo{author}{Bertran~Ferrer, M.}, \bibinfo{year}{2019}.
\newblock \bibinfo{title}{Bat recordings split data set}.
\newblock \bibinfo{journal}{Zenodo} \DOIprefix\doi{10.1038/s41598-023-49989-z}.
\bibitem[{Bouza et~al.(2023)Bouza, Bugeau and Lannelongue}]{2023BouzaBL}
\bibinfo{author}{Bouza, L.}, \bibinfo{author}{Bugeau, A.},
  \bibinfo{author}{Lannelongue, L.}, \bibinfo{year}{2023}.
\newblock \bibinfo{title}{How to estimate carbon footprint when training deep
  learning models? a guide and review}.
\newblock \bibinfo{journal}{Environmental Research Communications}
  \bibinfo{volume}{5}, \bibinfo{pages}{115014}.
\newblock \URLprefix \url{https://dx.doi.org/10.1088/2515-7620/acf81b},
  \DOIprefix\doi{10.1088/2515-7620/acf81b}.
\bibitem[{Bradfer-Lawrence et~al.(2023)Bradfer-Lawrence, Desjonqueres,
  Eldridge, Johnston and Metcalf}]{2023BradferDEJM}
\bibinfo{author}{Bradfer-Lawrence, T.}, \bibinfo{author}{Desjonqueres, C.},
  \bibinfo{author}{Eldridge, A.}, \bibinfo{author}{Johnston, A.},
  \bibinfo{author}{Metcalf, O.}, \bibinfo{year}{2023}.
\newblock \bibinfo{title}{Using acoustic indices in ecology: Guidance on study
  design, analyses and interpretation}.
\newblock \bibinfo{journal}{Methods in Ecology and Evolution}
  \bibinfo{volume}{14 (9)}, \bibinfo{pages}{2192--2204}.
\newblock \DOIprefix\doi{10.1111/2041-210X.14194}.
\bibitem[{Brownell(1997)}]{1997Brownell}
\bibinfo{author}{Brownell, W.}, \bibinfo{year}{1997}.
\newblock \bibinfo{title}{How the ear works - nature's solutions for
  listening}.
\newblock \bibinfo{journal}{Volta Rev} \bibinfo{volume}{99 (5)}.
\newblock \DOIprefix\doi{PMC2888317}.
\bibitem[{Catto et~al.(2003)Catto, Coyte, Agate and Langton}]{2003CattoCAL}
\bibinfo{author}{Catto, C.}, \bibinfo{author}{Coyte, A.},
  \bibinfo{author}{Agate, J.}, \bibinfo{author}{Langton, S.},
  \bibinfo{year}{2003}.
\newblock \bibinfo{title}{Bats as Indicators of Environmental Quality R\&D
  Technical Report E1-129/TR}.
\newblock \bibinfo{type}{Technical Report} \bibinfo{number}{ISBN: 1 844 32251
  3}. Environment Agency. \bibinfo{address}{Rio House, Waterside Drive, Aztec
  West, Almondsbury Bristol BS32 4UD}.
\newblock \bibinfo{note}{N.B. This document was produced under R\&D Project
  E1-129 by the Bat Conservation Trust.}
\bibitem[{Chen et~al.(2019)Chen, Li, Tao, Barnett, Su and Rudin}]{2019ChenLTB}
\bibinfo{author}{Chen, C.}, \bibinfo{author}{Li, O.}, \bibinfo{author}{Tao,
  C.}, \bibinfo{author}{Barnett, A.}, \bibinfo{author}{Su, J.},
  \bibinfo{author}{Rudin, C.}, \bibinfo{year}{2019}.
\newblock \bibinfo{title}{This looks like that: Deep learning for interpretable
  image recognition}.
\newblock \URLprefix \url{https://arxiv.org/abs/1806.10574},
  \href{http://arxiv.org/abs/1806.10574}{{\tt arXiv:1806.10574}}.
\bibitem[{Chen and Aihara(1995)}]{1995ChenA}
\bibinfo{author}{Chen, L.}, \bibinfo{author}{Aihara, K.}, \bibinfo{year}{1995}.
\newblock \bibinfo{title}{Chaotic simulated annealing by a neural network model
  with transient chaos}.
\newblock \bibinfo{journal}{Neural Networks} \bibinfo{volume}{8},
  \bibinfo{pages}{915--930}.
\newblock \DOIprefix\doi{10.1016/0893-6080(95)00033-V}.
\bibitem[{Dai and Nakano(1998)}]{1998DaiN}
\bibinfo{author}{Dai, Y.}, \bibinfo{author}{Nakano, Y.}, \bibinfo{year}{1998}.
\newblock \bibinfo{title}{Recognition of facial images with low resolution
  using a hopfield memory model}.
\newblock \bibinfo{journal}{Pattern Recognition} \bibinfo{volume}{31 (2)},
  \bibinfo{pages}{159--167}.
\newblock \DOIprefix\doi{10.1016/S0031-3203(97)00040-X.}
\bibitem[{Dobie and Van~Hemel(2004)}]{2004DobieH}
\bibinfo{author}{Dobie, R.}, \bibinfo{author}{Van~Hemel, S.},
  \bibinfo{year}{2004}.
\newblock \bibinfo{title}{Hearing Loss: Determining Eligibility for Social
  Security Benefits}. \bibinfo{publisher}{National Academies Press (US)}.
  chapter~\bibinfo{chapter}{2}.
\newblock pp. \bibinfo{pages}{42--68}.
\newblock \DOIprefix\doi{10.17226/11099}.
\bibitem[{Dosovitskiy et~al.(2020)Dosovitskiy, Beyer, Kolesnikov, Weissenborn,
  Zhai, Unterthiner, Dehghani, Minderer, Heigold, Gelly, Uszkoreit and
  Houlsby}]{2020DosovitskiyBKW}
\bibinfo{author}{Dosovitskiy, A.}, \bibinfo{author}{Beyer, L.},
  \bibinfo{author}{Kolesnikov, A.}, \bibinfo{author}{Weissenborn, D.},
  \bibinfo{author}{Zhai, X.}, \bibinfo{author}{Unterthiner, T.},
  \bibinfo{author}{Dehghani, M.}, \bibinfo{author}{Minderer, M.},
  \bibinfo{author}{Heigold, G.}, \bibinfo{author}{Gelly, S.},
  \bibinfo{author}{Uszkoreit, J.}, \bibinfo{author}{Houlsby, N.},
  \bibinfo{year}{2020}.
\newblock \bibinfo{title}{An image is worth 16x16 words: Transformers for image
  recognition at scale}.
\newblock \bibinfo{journal}{CoRR} \bibinfo{volume}{abs/2010.11929}.
\newblock \URLprefix \url{https://arxiv.org/abs/2010.11929}.
\bibitem[{Dufourq et~al.(2022)Dufourq, Batist, Foquet and
  Durbach}]{2022DufourqBFD}
\bibinfo{author}{Dufourq, E.}, \bibinfo{author}{Batist, C.},
  \bibinfo{author}{Foquet, R.}, \bibinfo{author}{Durbach, I.},
  \bibinfo{year}{2022}.
\newblock \bibinfo{title}{Passive acoustic monitoring of animal populations
  with transfer learning}.
\newblock \bibinfo{journal}{Ecological Informatics} \bibinfo{volume}{70},
  \bibinfo{pages}{101688}.
\newblock \DOIprefix\doi{10.1016/j.ecoinf.2022.101688}.
\bibitem[{Ferreira et~al.(2025)Ferreira, Felipe~{da Silva}, Mesquita, Rosa,
  Buchmann and Mesquita-Neto}]{2025FerreiraFMR}
\bibinfo{author}{Ferreira, A.}, \bibinfo{author}{Felipe~{da Silva}, N.},
  \bibinfo{author}{Mesquita, F.}, \bibinfo{author}{Rosa, T.},
  \bibinfo{author}{Buchmann, S.}, \bibinfo{author}{Mesquita-Neto, J.},
  \bibinfo{year}{2025}.
\newblock \bibinfo{title}{Transformer models improve the acoustic recognition
  of buzz-pollinating bee species}.
\newblock \bibinfo{journal}{Ecological Informatics} \bibinfo{volume}{86},
  \bibinfo{pages}{103010}.
\newblock \DOIprefix\doi{10.1016/j.ecoinf.2025.103010}.
\bibitem[{Funosas et~al.(2024)Funosas, Barbaro, Schillé, Elger, Castagneyrol
  and Cauchoix}]{2024FunosasBSE}
\bibinfo{author}{Funosas, D.}, \bibinfo{author}{Barbaro, L.},
  \bibinfo{author}{Schillé, L.}, \bibinfo{author}{Elger, A.},
  \bibinfo{author}{Castagneyrol, B.}, \bibinfo{author}{Cauchoix, M.},
  \bibinfo{year}{2024}.
\newblock \bibinfo{title}{Assessing the potential of birdnet to infer european
  bird communities from large-scale ecoacoustic data}.
\newblock \bibinfo{journal}{Ecological Indicators} \bibinfo{volume}{164},
  \bibinfo{pages}{112146}.
\newblock \DOIprefix\doi{10.1016/j.ecolind.2024.112146}.
\bibitem[{Ghani et~al.(2023)Ghani, Denton, Kahl and Klinck}]{2023GhaniDKK}
\bibinfo{author}{Ghani, B.}, \bibinfo{author}{Denton, T.},
  \bibinfo{author}{Kahl, S.}, \bibinfo{author}{Klinck, H.},
  \bibinfo{year}{2023}.
\newblock \bibinfo{title}{Global birdsong embeddings enable superior transfer
  learning for bioacoustic classification}.
\newblock \bibinfo{journal}{Scientific Reports} \bibinfo{volume}{13},
  \bibinfo{pages}{22876}.
\newblock \DOIprefix\doi{10.1038/s41598-023-49989-z}.
\bibitem[{Goodwin and Gillam(2021)}]{2021GoodwinG}
\bibinfo{author}{Goodwin, K.}, \bibinfo{author}{Gillam, E.},
  \bibinfo{year}{2021}.
\newblock \bibinfo{title}{Testing accuracy and agreement among multiple
  versions of automated bat call classification software}.
\newblock \bibinfo{journal}{Wildlife Society Bulletin} \bibinfo{volume}{45},
  \bibinfo{pages}{690--705}.
\newblock \DOIprefix\doi{10.1002/wsb.1235}.
\bibitem[{Hebb(1949)}]{1949Hebb}
\bibinfo{author}{Hebb, D.}, \bibinfo{year}{1949}.
\newblock \bibinfo{title}{The Organization of Behavior: A Neuropsychological
  Theory, 1st ed.}
\newblock \bibinfo{publisher}{Wiley}.
\bibitem[{Heinrich et~al.(2025)Heinrich, Rauch, Sick and
  Scholz}]{2025HeinrichRSS}
\bibinfo{author}{Heinrich, R.}, \bibinfo{author}{Rauch, L.},
  \bibinfo{author}{Sick, B.}, \bibinfo{author}{Scholz, C.},
  \bibinfo{year}{2025}.
\newblock \bibinfo{title}{Audioprotopnet: An interpretable deep learning model
  for bird sound classification}.
\newblock \bibinfo{journal}{Ecological Informatics} \bibinfo{volume}{87},
  \bibinfo{pages}{103081}.
\newblock \DOIprefix\doi{10.1016/j.ecoinf.2025.103081}.
\bibitem[{Hertz(1991)}]{1991Hertz}
\bibinfo{author}{Hertz, J.}, \bibinfo{year}{1991}.
\newblock \bibinfo{title}{Introduction To The Theory Of Neural Computation (1st
  ed.)}.
\newblock \bibinfo{publisher}{CRC Press}.
\newblock \DOIprefix\doi{10.1201/9780429499661}.
\bibitem[{Hill et~al.(2019)Hill, Prince, Snaddon, Doncaster and
  Rogers}]{2019HillPSD}
\bibinfo{author}{Hill, A.}, \bibinfo{author}{Prince, P.},
  \bibinfo{author}{Snaddon, J.}, \bibinfo{author}{Doncaster, C.},
  \bibinfo{author}{Rogers, A.}, \bibinfo{year}{2019}.
\newblock \bibinfo{title}{Audiomoth: A low-cost acoustic device for monitoring
  biodiversity and the environment}.
\newblock \bibinfo{journal}{HardwareX} \bibinfo{volume}{6},
  \bibinfo{pages}{e00073}.
\newblock \DOIprefix\doi{10.1016/j.ohx.2019.e00073}.
\bibitem[{Hopfield(1982)}]{1982Hopfield}
\bibinfo{author}{Hopfield, J.}, \bibinfo{year}{1982}.
\newblock \bibinfo{title}{Neural networks and physical systems with emergent
  collective computational abilities}.
\newblock \bibinfo{journal}{Proceedings of the National Academy of Sciences,
  USA} \bibinfo{volume}{79}, \bibinfo{pages}{2554--2558}.
\newblock \DOIprefix\doi{10.1073/pnas.79.8.2554}.
\bibitem[{Hopfield(1984)}]{1984Hopfield}
\bibinfo{author}{Hopfield, J.}, \bibinfo{year}{1984}.
\newblock \bibinfo{title}{Neurons with graded response have collective
  computational properties like those of two-state neurons}.
\newblock \bibinfo{journal}{Proceedings of the National Academy of Sciences of
  the United States of America} \bibinfo{volume}{81 (10)},
  \bibinfo{pages}{3088--3092}.
\newblock \DOIprefix\doi{10.1073/pnas.81.10.3088}.
\bibitem[{Hopfield and Tank(1985)}]{1985HopfieldT}
\bibinfo{author}{Hopfield, J.}, \bibinfo{author}{Tank, D.},
  \bibinfo{year}{1985}.
\newblock \bibinfo{title}{"neural" computation of decisions in optimization
  problems}.
\newblock \bibinfo{journal}{Biological cybernetics} \bibinfo{volume}{52},
  \bibinfo{pages}{141--152}.
\newblock \DOIprefix\doi{10.1007/BF00339943}.
\bibitem[{Kahl et~al.(2021)Kahl, Wood, Eibl and Klinck}]{2021KahlWEK}
\bibinfo{author}{Kahl, S.}, \bibinfo{author}{Wood, C.}, \bibinfo{author}{Eibl,
  M.}, \bibinfo{author}{Klinck, H.}, \bibinfo{year}{2021}.
\newblock \bibinfo{title}{Birdnet: A deep learning solution for avian diversity
  monitoring}.
\newblock \bibinfo{journal}{Ecological Informatics} \bibinfo{volume}{61},
  \bibinfo{pages}{101236}.
\newblock \DOIprefix\doi{10.1016/j.ecoinf.2021.101236}.
\bibitem[{Kershenbaum et~al.(2025)Kershenbaum, Akçay, Babu-Saheer, Barnhill,
  Best, Cauzinille, Clink, Dassow, Dufourq, Growcott, Markham, Marti-Domken,
  Marxer, Muir, Reynolds, Root-Gutteridge, Sadhukhan, Schindler, Smith,
  Stowell, Wascher and Dunn}]{2025KershenbaumABB}
\bibinfo{author}{Kershenbaum, A.}, \bibinfo{author}{Akçay, {\c{C}}.},
  \bibinfo{author}{Babu-Saheer, L.}, \bibinfo{author}{Barnhill, A.},
  \bibinfo{author}{Best, P.}, \bibinfo{author}{Cauzinille, J.},
  \bibinfo{author}{Clink, D.}, \bibinfo{author}{Dassow, A.},
  \bibinfo{author}{Dufourq, E.}, \bibinfo{author}{Growcott, J.},
  \bibinfo{author}{Markham, A.}, \bibinfo{author}{Marti-Domken, B.},
  \bibinfo{author}{Marxer, R.}, \bibinfo{author}{Muir, J.},
  \bibinfo{author}{Reynolds, S.}, \bibinfo{author}{Root-Gutteridge, H.},
  \bibinfo{author}{Sadhukhan, S.}, \bibinfo{author}{Schindler, L.},
  \bibinfo{author}{Smith, B.}, \bibinfo{author}{Stowell, D.},
  \bibinfo{author}{Wascher, C.}, \bibinfo{author}{Dunn, J.},
  \bibinfo{year}{2025}.
\newblock \bibinfo{title}{Automatic detection for bioacoustic research: a
  practical guide from and for biologists and computer scientists}.
\newblock \bibinfo{journal}{Biological Reviews} \bibinfo{volume}{100},
  \bibinfo{pages}{620--646}.
\newblock \DOIprefix\doi{10.1111/brv.13155}.
\bibitem[{Liu et~al.(2023)Liu, Gao, Chen, Zhou, Peng, Yu, Ma and
  Wang}]{2023LiuGCZ}
\bibinfo{author}{Liu, S.}, \bibinfo{author}{Gao, X.}, \bibinfo{author}{Chen,
  L.}, \bibinfo{author}{Zhou, S.}, \bibinfo{author}{Peng, Y.},
  \bibinfo{author}{Yu, D.}, \bibinfo{author}{Ma, X.}, \bibinfo{author}{Wang,
  Y.}, \bibinfo{year}{2023}.
\newblock \bibinfo{title}{Multi-traveler salesman problem for unmanned
  vehicles: Optimization through improved hopfield neural network}.
\newblock \bibinfo{journal}{Sustainability} \bibinfo{volume}{15}.
\newblock \DOIprefix\doi{10.3390/su152015118}.
\bibitem[{Liu et~al.(2022)Liu, Mao, Wu, Feichtenhofer, Darrell and
  Xie}]{2022LiuMWF}
\bibinfo{author}{Liu, Z.}, \bibinfo{author}{Mao, H.}, \bibinfo{author}{Wu, C.},
  \bibinfo{author}{Feichtenhofer, C.}, \bibinfo{author}{Darrell, T.},
  \bibinfo{author}{Xie, S.}, \bibinfo{year}{2022}.
\newblock \bibinfo{title}{A convnet for the 2020s}.
\newblock \URLprefix \url{https://arxiv.org/abs/2201.03545},
  \href{http://arxiv.org/abs/2201.03545}{{\tt arXiv:2201.03545}}.
\bibitem[{Mac~Aodha et~al.(2018)Mac~Aodha, Gibb, Barlow, Browning, Firman and
  Freeman~et al}]{2018MacGBBFF}
\bibinfo{author}{Mac~Aodha, O.}, \bibinfo{author}{Gibb, R.},
  \bibinfo{author}{Barlow, K.}, \bibinfo{author}{Browning, E.},
  \bibinfo{author}{Firman, M.}, \bibinfo{author}{Freeman~et al, R.},
  \bibinfo{year}{2018}.
\newblock \bibinfo{title}{Bat detective—deep learning tools for bat acoustic
  signal detection}.
\newblock \bibinfo{journal}{PLOS Computational Biology} \bibinfo{volume}{14
  (3)}, \bibinfo{pages}{e1005995}.
\newblock \DOIprefix\doi{10.1371/journal.pcbi.1005995}.
\bibitem[{MacIsaac et~al.(2024)MacIsaac, Newson, Ashton-Butt, Pearce and
  Milner}]{2024MacIsaacNAP}
\bibinfo{author}{MacIsaac, J.}, \bibinfo{author}{Newson, S.},
  \bibinfo{author}{Ashton-Butt, A.}, \bibinfo{author}{Pearce, H.},
  \bibinfo{author}{Milner, B.}, \bibinfo{year}{2024}.
\newblock \bibinfo{title}{Improving acoustic species identification using data
  augmentation within a deep learning framework}.
\newblock \bibinfo{journal}{Ecological Informatics} \bibinfo{volume}{83},
  \bibinfo{pages}{102851}.
\newblock \DOIprefix\doi{10.1016/j.ecoinf.2024.102851}.
\bibitem[{Manzano-Rubio et~al.(2022)Manzano-Rubio, Bota, Brotons, Soto-Largo
  and Pérez-Granados}]{2022ManzanoBBSP}
\bibinfo{author}{Manzano-Rubio, R.}, \bibinfo{author}{Bota, G.},
  \bibinfo{author}{Brotons, L.}, \bibinfo{author}{Soto-Largo, E.},
  \bibinfo{author}{Pérez-Granados, C.}, \bibinfo{year}{2022}.
\newblock \bibinfo{title}{Low-cost open-source recorders and ready-to-use
  machine learning approaches provide effective monitoring of threatened
  species}.
\newblock \bibinfo{journal}{Ecological Informatics} \bibinfo{volume}{72},
  \bibinfo{pages}{101910}.
\newblock \DOIprefix\doi{10.1016/j.ecoinf.2022.101910}.
\bibitem[{Marchal et~al.(2022)Marchal, Fabianek and Aubry}]{2022MarchalFA}
\bibinfo{author}{Marchal, J.}, \bibinfo{author}{Fabianek, F.},
  \bibinfo{author}{Aubry, Y.}, \bibinfo{year}{2022}.
\newblock \bibinfo{title}{Software performance for the automated identification
  of bird vocalisations: the case of two closely related species}.
\newblock \bibinfo{journal}{Bioacoustics} \bibinfo{volume}{31 (4)},
  \bibinfo{pages}{397--413}.
\newblock \DOIprefix\doi{10.1080/09524622.2021.1945952}.
\bibitem[{McEwen et~al.(2024)McEwen, Soltero, Gutschmidt, Bainbridge-Smith,
  Atlas and Green}]{2024McEwanKGB}
\bibinfo{author}{McEwen, B.}, \bibinfo{author}{Soltero, K.},
  \bibinfo{author}{Gutschmidt, S.}, \bibinfo{author}{Bainbridge-Smith, A.},
  \bibinfo{author}{Atlas, J.}, \bibinfo{author}{Green, R.},
  \bibinfo{year}{2024}.
\newblock \bibinfo{title}{Active few-shot learning for rare bioacoustic feature
  annotation}.
\newblock \bibinfo{journal}{Ecological Informatics} \bibinfo{volume}{82},
  \bibinfo{pages}{102734}.
\newblock \DOIprefix\doi{10.1016/j.ecoinf.2024.102734}.
\bibitem[{Metcalf et~al.(2023)Metcalf, Abrahams, Ashington, Baker,
  Bradfer-Lawrence, Browning, Carruthers-Jones, Darby, Dick, Eldridge, Elliott,
  Heath, Howden-Leach, Johnston, Lees, Meyer, Ruiz~Arana and
  Smyth}]{2023MetcalfAAB}
\bibinfo{author}{Metcalf, O.}, \bibinfo{author}{Abrahams, C.},
  \bibinfo{author}{Ashington, B.}, \bibinfo{author}{Baker, E.},
  \bibinfo{author}{Bradfer-Lawrence, T.}, \bibinfo{author}{Browning, E.},
  \bibinfo{author}{Carruthers-Jones, J.}, \bibinfo{author}{Darby, J.},
  \bibinfo{author}{Dick, J.}, \bibinfo{author}{Eldridge, A.},
  \bibinfo{author}{Elliott, D.}, \bibinfo{author}{Heath, B.},
  \bibinfo{author}{Howden-Leach, P.}, \bibinfo{author}{Johnston, A.},
  \bibinfo{author}{Lees, A.}, \bibinfo{author}{Meyer, C.},
  \bibinfo{author}{Ruiz~Arana, U.}, \bibinfo{author}{Smyth, S.},
  \bibinfo{year}{2023}.
\newblock \bibinfo{title}{Good practice guidelines for long-term ecoacoustic
  monitoring in the UK}.
\newblock \bibinfo{type}{Technical Report}. The UK Acoustics Network.
\newblock \URLprefix
  \url{https://e-space.mmu.ac.uk/631466/1/Good_practice_guidelines_for_long-term%20%281%29.pdf}.
\bibitem[{Mydlarz et~al.(2017)Mydlarz, Salamon and Bello}]{2017MydlarzSB}
\bibinfo{author}{Mydlarz, C.}, \bibinfo{author}{Salamon, J.},
  \bibinfo{author}{Bello, J.}, \bibinfo{year}{2017}.
\newblock \bibinfo{title}{The implementation of low-cost urban acoustic
  monitoring devices}.
\newblock \bibinfo{journal}{Applied Acoustics} \bibinfo{volume}{117},
  \bibinfo{pages}{207--218}.
\newblock \DOIprefix\doi{10.1016/j.apacoust.2016.06.010}.
  \bibinfo{note}{acoustics in Smart Cities}.
\bibitem[{{NobelPrize.org}(2024)}]{2024Nobel}
\bibinfo{author}{{NobelPrize.org}}, \bibinfo{year}{2024}.
\newblock \bibinfo{title}{The nobel prize in physics 2024}.
\newblock
  \bibinfo{howpublished}{https://www.nobelprize.org/uploads/2024/11/press-physicsprize2024-2.pdf}.
\newblock \bibinfo{note}{Accessed: 2025-04-04}.
\bibitem[{Northcutt et~al.(2021)Northcutt, Athalye and
  Mueller}]{2021NorthcuttAM}
\bibinfo{author}{Northcutt, C.}, \bibinfo{author}{Athalye, A.},
  \bibinfo{author}{Mueller, J.}, \bibinfo{year}{2021}.
\newblock \bibinfo{title}{Pervasive label errors in test sets destabilize
  machine learning benchmarks}.
\newblock \URLprefix \url{https://arxiv.org/abs/2103.14749},
  \href{http://arxiv.org/abs/2103.14749}{{\tt arXiv:2103.14749}}.
\bibitem[{Pierce and Griffin(1938)}]{1938PierceG}
\bibinfo{author}{Pierce, G.}, \bibinfo{author}{Griffin, D.},
  \bibinfo{year}{1938}.
\newblock \bibinfo{title}{Experimental determination of supersonic notes
  emitted by bats}.
\newblock \bibinfo{journal}{Journal of Mammalogy} \bibinfo{volume}{19 (4)},
  \bibinfo{pages}{454--455}.
\newblock \DOIprefix\doi{10.2307/1374231}.
\bibitem[{{Planque, B. }(2024)}]{2024Planque}
\bibinfo{author}{{Planque, B. }}, \bibinfo{year}{2024}.
\newblock \bibinfo{title}{A short introduction to machine learning results}.
\newblock \bibinfo{howpublished}{https://xeno-canto.org/article/299}.
\newblock \bibinfo{note}{Accessed: 2025-04-04}.
\bibitem[{Pérez-Granados(2023)}]{2023Perez}
\bibinfo{author}{Pérez-Granados, C.}, \bibinfo{year}{2023}.
\newblock \bibinfo{title}{Birdnet: applications, performance, pitfalls and
  future opportunities}.
\newblock \bibinfo{journal}{Ibis} \bibinfo{volume}{165},
  \bibinfo{pages}{1068--1075}.
\newblock \DOIprefix\doi{10.1111/ibi.13193}.
\bibitem[{Pérez-Granados and Traba(2021)}]{2021PerezT}
\bibinfo{author}{Pérez-Granados, C.}, \bibinfo{author}{Traba, J.},
  \bibinfo{year}{2021}.
\newblock \bibinfo{title}{Estimating bird density using passive acoustic
  monitoring: a review of methods and suggestions for further research}.
\newblock \bibinfo{journal}{Ibis} \bibinfo{volume}{163},
  \bibinfo{pages}{765--783}.
\newblock \DOIprefix\doi{10.1111/ibi.12944}.
\bibitem[{Rasmussen et~al.(2024)Rasmussen, Stowell and
  Briefer}]{2024RasmussenSB}
\bibinfo{author}{Rasmussen, J.}, \bibinfo{author}{Stowell, D.},
  \bibinfo{author}{Briefer, E.}, \bibinfo{year}{2024}.
\newblock \bibinfo{title}{Sound evidence for biodiversity monitoring}.
\newblock \bibinfo{journal}{Science} \bibinfo{volume}{385},
  \bibinfo{pages}{138--140}.
\newblock \DOIprefix\doi{10.1126/science.adh2716}.
\bibitem[{Rodden et~al.(2024)Rodden, Gascoyne, Naughton, Brennan and
  Parkes}]{2024RoddenGNBP}
\bibinfo{author}{Rodden, E.}, \bibinfo{author}{Gascoyne, A.},
  \bibinfo{author}{Naughton, L.}, \bibinfo{author}{Brennan, J.},
  \bibinfo{author}{Parkes, A.}, \bibinfo{year}{2024}.
\newblock \bibinfo{title}{Transient chaotic neural network with negative
  self-feedback memory for continuous optimisation problems}, in:
  \bibinfo{editor}{Arai, K.} (Ed.), \bibinfo{booktitle}{Proceedings of the
  Future Technologies Conference (FTC) 2024, Volume 1},
  \bibinfo{publisher}{Springer Nature Switzerland}, \bibinfo{address}{Cham}.
  pp. \bibinfo{pages}{290--303}.
\bibitem[{Russ(2021)}]{2021Russ}
\bibinfo{author}{Russ, J.}, \bibinfo{year}{2021}.
\newblock \bibinfo{title}{Bat calls of Britain and Europe}.
\newblock Bat Biology and Conservation, \bibinfo{publisher}{Pelagic
  Publishing}, \bibinfo{address}{Exeter, England}.
\bibitem[{Salem et~al.(2024)Salem, Shirayama, Shimazaki and Oki}]{2024SalemSSO}
\bibinfo{author}{Salem, S.}, \bibinfo{author}{Shirayama, S.},
  \bibinfo{author}{Shimazaki, S.}, \bibinfo{author}{Oki, K.},
  \bibinfo{year}{2024}.
\newblock \bibinfo{title}{Ensemble deep learning and anomaly detection
  framework for automatic audio classification: Insights into deer
  vocalizations}.
\newblock \bibinfo{journal}{Ecological Informatics} \bibinfo{volume}{84},
  \bibinfo{pages}{102883}.
\newblock \DOIprefix\doi{10.1016/j.ecoinf.2024.102883}.
\bibitem[{Sandbrook(2025)}]{2025Sandbrook}
\bibinfo{author}{Sandbrook, C.}, \bibinfo{year}{2025}.
\newblock \bibinfo{title}{Beyond the hype: Navigating the conservation
  implications of artificial intelligence}.
\newblock \bibinfo{journal}{Conservation Letters} \bibinfo{volume}{18},
  \bibinfo{pages}{e13076}.
\newblock \DOIprefix\doi{10.1111/conl.13076}.
\bibitem[{Sethi et~al.(2022)Sethi, Kovac, Wiesemüller, Miriyev and
  Boutry}]{2022SethiKWM}
\bibinfo{author}{Sethi, S.}, \bibinfo{author}{Kovac, M.},
  \bibinfo{author}{Wiesemüller, F.}, \bibinfo{author}{Miriyev, A.},
  \bibinfo{author}{Boutry, C.}, \bibinfo{year}{2022}.
\newblock \bibinfo{title}{Biodegradable sensors are ready to transform
  autonomous ecological monitoring}.
\newblock \bibinfo{journal}{Nature Ecology and Evolution} \bibinfo{volume}{6},
  \bibinfo{pages}{1245--1247}.
\newblock \DOIprefix\doi{10.1038/s41559-022-01824-w}.
\bibitem[{Stowell(2022)}]{2022Stowell}
\bibinfo{author}{Stowell, D.}, \bibinfo{year}{2022}.
\newblock \bibinfo{title}{Computational bioacoustics with deep learning: a
  review and roadmap.}
\newblock \bibinfo{journal}{PeerJ} \bibinfo{volume}{10},
  \bibinfo{pages}{e13152}.
\newblock \DOIprefix\doi{10.7717/peerj.13152}.
\bibitem[{Szesciorka et~al.(2023)Szesciorka, McCullough and
  Oleson}]{2023SzesciorkaMO}
\bibinfo{author}{Szesciorka, A.}, \bibinfo{author}{McCullough, L.K.},
  \bibinfo{author}{Oleson, E.}, \bibinfo{year}{2023}.
\newblock \bibinfo{title}{An unknown nocturnal call type in the mariana
  archipelago}.
\newblock \bibinfo{journal}{JASA Express Lett} \bibinfo{volume}{3 (1)},
  \bibinfo{pages}{011201}.
\newblock \DOIprefix\doi{10.1121/10.0017068}.
\bibitem[{Tabak et~al.(2022)Tabak, Murray, Reed, Lombardi and
  Bay}]{2022TabakMRLB}
\bibinfo{author}{Tabak, M.}, \bibinfo{author}{Murray, K.},
  \bibinfo{author}{Reed, A.}, \bibinfo{author}{Lombardi, J.},
  \bibinfo{author}{Bay, K.}, \bibinfo{year}{2022}.
\newblock \bibinfo{title}{Automated classification of bat echolocation call
  recordings with artificial intelligence}.
\newblock \bibinfo{journal}{Ecological Informatics} \bibinfo{volume}{68},
  \bibinfo{pages}{101526}.
\newblock \DOIprefix\doi{10.1016/j.ecoinf.2021.101526}.
\bibitem[{Teixeira et~al.(2019)Teixeira, Maron and van
  Rensburg}]{2019TeixeiraMR}
\bibinfo{author}{Teixeira, D.}, \bibinfo{author}{Maron, M.},
  \bibinfo{author}{van Rensburg, B.}, \bibinfo{year}{2019}.
\newblock \bibinfo{title}{Bioacoustic monitoring of animal vocal behavior for
  conservation}.
\newblock \bibinfo{journal}{Conservation Science and Practice}
  \bibinfo{volume}{1 (8)}.
\newblock \DOIprefix\doi{10.1111/csp2.72}.
\bibitem[{Thompson et~al.(2021)Thompson, Greenewald, Lee and
  Manso}]{2021ThompsonGLM}
\bibinfo{author}{Thompson, N.C.}, \bibinfo{author}{Greenewald, K.},
  \bibinfo{author}{Lee, K.}, \bibinfo{author}{Manso, G.F.},
  \bibinfo{year}{2021}.
\newblock \bibinfo{title}{Deep learning's diminishing returns: The cost of
  improvement is becoming unsustainable}.
\newblock \bibinfo{journal}{IEEE Spectrum} \bibinfo{volume}{58},
  \bibinfo{pages}{50--55}.
\newblock \DOIprefix\doi{10.1109/MSPEC.2021.9563954}.
\bibitem[{{Toews, R. }(2020)}]{2020Toews}
\bibinfo{author}{{Toews, R. }}, \bibinfo{year}{2020}.
\newblock \bibinfo{title}{Deep learning’s carbon emissions problem}.
\newblock
  \bibinfo{howpublished}{https://www.forbes.com/sites/robtoews/2020/06/17/deep-learnings-climate-change-problem/}.
\newblock \bibinfo{note}{Accessed: 2025-04-04}.
\bibitem[{{Wildlife Acoustics}(2017)}]{2017WA}
\bibinfo{author}{{Wildlife Acoustics}}, \bibinfo{year}{2017}.
\newblock \bibinfo{title}{Kaleidoscope pro bat auto id how it works}.
\newblock \bibinfo{howpublished}{https://vimeo.com/182997396}.
\newblock \bibinfo{note}{Accessed: 2025-04-04}.
\bibitem[{{Wildlife Acoustics}(2024)}]{2024WA}
\bibinfo{author}{{Wildlife Acoustics}}, \bibinfo{year}{2024}.
\newblock \bibinfo{title}{Products}.
\newblock \bibinfo{howpublished}{https://www.wildlifeacoustics.com/products}.
\newblock \bibinfo{note}{Accessed: 2025-04-04}.
\bibitem[{Yoh et~al.(2022)Yoh, Kingston, McArthur, Aylen, Huang, Jinggong,
  Khan, Lee, Mitchell, Bicknell and Struebig}]{2022YohKMA}
\bibinfo{author}{Yoh, N.}, \bibinfo{author}{Kingston, T.},
  \bibinfo{author}{McArthur, E.}, \bibinfo{author}{Aylen, O.},
  \bibinfo{author}{Huang, J.}, \bibinfo{author}{Jinggong, E.},
  \bibinfo{author}{Khan, F.}, \bibinfo{author}{Lee, B.},
  \bibinfo{author}{Mitchell, S.}, \bibinfo{author}{Bicknell, J.},
  \bibinfo{author}{Struebig, M.}, \bibinfo{year}{2022}.
\newblock \bibinfo{title}{A machine learning framework to classify southeast
  asian echolocating bats}.
\newblock \bibinfo{journal}{Ecological Indicators} \bibinfo{volume}{136},
  \bibinfo{pages}{108696}.
\newblock \DOIprefix\doi{10.1016/j.ecolind.2022.108696}.
\bibitem[{You et~al.(2018)You, Zhang, Hsieh, Demmel and Keutzer}]{2018YouZHDK}
\bibinfo{author}{You, Y.}, \bibinfo{author}{Zhang, Z.}, \bibinfo{author}{Hsieh,
  C.J.}, \bibinfo{author}{Demmel, J.}, \bibinfo{author}{Keutzer, K.},
  \bibinfo{year}{2018}.
\newblock \bibinfo{title}{Imagenet training in minutes}.
\newblock \URLprefix \url{https://arxiv.org/abs/1709.05011},
  \href{http://arxiv.org/abs/1709.05011}{{\tt arXiv:1709.05011}}.

\end{thebibliography}







\end{document}